\documentclass[letterpaper, 10 pt, conference]{ieeeconf}
\IEEEoverridecommandlockouts
\usepackage{graphicx} 
\usepackage{epsfig}
\usepackage{amsmath}
\usepackage{array}
\usepackage{amssymb}
\usepackage{cite}
\usepackage{multirow}
\usepackage{subcaption}
\usepackage{booktabs}
\usepackage{mathtools}
\usepackage{siunitx}
\usepackage{comment}
\usepackage{threeparttable}
\usepackage{bm}
\usepackage{pifont}
\usepackage{color}
\usepackage{soul}
\usepackage{algorithmicx}
\usepackage[ruled]{algorithm}
\usepackage[noend]{algpseudocode}
\usepackage{cleveref}

\makeatletter
\newcommand{\subsubsubsection}{\@startsection{paragraph}{4}{\z@}%
  {1.0\Cvs \@plus.5\Cdp \@minus.2\Cdp}%
  {.1\Cvs \@plus.3\Cdp}%
  {\reset@font\sffamily\normalsize}
}
\makeatother
\newcommand{\figlab}[1]{\label{fig:#1}}
\newcommand{\figref}[1]{Fig.~\ref{fig:#1}} 
\newcommand{\tablab}[1]{\label{tab:#1}}
\newcommand{\tabref}[1]{Table~\ref{tab:#1}} 
\newcommand{\forlab}[1]{\label{for:#1}}
\newcommand{\forref}[1]{Equation~(\ref{for:#1})} 

\newcommand{\etal}{\textit{et al.}}
\newcommand{\ie}{\textit{i.e.}}
\newcommand{\eg}{\textit{e.g.}}

\begin{document}

\title{Affordance-Guided Enveloping Grasp Demonstration \\Toward Non-destructive Disassembly of Pinch-Infeasible Mating Parts}

\author{Masaki Tsutsumi$^{1}$, Takuya Kiyokawa$^{1}$, Gen Sako$^{1}$, and Kensuke Harada$^{1,2}$%
\thanks{$^{1}$Department of Systems Innovation, Graduate School of Engineering Science, The University of Osaka, 1-3 Machikaneyama, Toyonaka, Osaka, Japan.}%
\thanks{$^{2}$Industrial Cyber-physical Systems Research Center, The National Institute of Advanced Industrial Science and Technology (AIST), 2-3-26 Aomi, Koto-ku, Tokyo, Japan.}%
}

\maketitle

\begin{abstract}
Robotic disassembly of complex mating components often renders pinch grasping infeasible, necessitating multi-fingered enveloping grasps. However, visual occlusions and geometric constraints complicate teaching appropriate grasp motions when relying solely on 2D camera feeds. To address this, we propose an affordance-guided teleoperation method that pre-generates enveloping grasp candidates via physics simulation. These Affordance Templates (ATs) are visualized with a color gradient reflecting grasp quality to augment operator perception. Simulations demonstrate the method's generality across various components. Real-robot experiments validate that AT-based visual augmentation enables operators to effectively select and teach enveloping grasp strategies for real-world disassembly, even under severe visual and geometric constraints.
\end{abstract}

\IEEEpeerreviewmaketitle

\section{Introduction}
The rapid accumulation of electronic waste (e-waste) has become a critical global environmental challenge. Non-destructive disassembly preserves high-value components for sustainable circulation, mitigating e-waste challenges more effectively than destructive recycling~\cite{Sundin2012,Heibeck2023}. Although automated disassembly planning is becoming feasible at a macroscopic level~\cite{Kiyokawa2025,Kiyokawa2026}, autonomous physical separation of tightly mated parts remains exceptionally difficult. Because industrial appliances are designed for rapid assembly with tight clearances and high-friction mating structures, autonomous extraction frequently fails due to unpredictable jamming or slipping. Therefore, keeping the human operator in the loop via semi-autonomous teleoperation is essential to manually modulate extraction forces and re-adjust finger placements when jamming or slipping occurs.

Furthermore, the lack of graspable protrusions on these mating components renders two-fingered pinch grasping physically infeasible. While multi-fingered hands capable of enveloping grasps are mechanically ideal for extracting such parts, teaching these high-DOF motions via conventional teleoperation relying on human spatial cognition is highly challenging. Because the target components are deeply embedded within the appliance casing, 2D cameras cannot provide the depth perception necessary to safely guide multiple fingers into narrow spaces without colliding with surrounding structures. Although affordance-based demonstration methods have been proposed~\cite{Gorjup2019,Regal2023,Sako2025}, they primarily assume simplified grasp models or parallel grippers, fundamentally limiting their applicability to multi-fingered enveloping grasps. Therefore, achieving both intuitive affordance presentation and high-DOF operability is crucial to reduce the operator's cognitive burden.

\begin{figure}[tb]
    \centering
    \begin{minipage}[tb]{\linewidth}
        \centering
        \includegraphics[keepaspectratio, width=\linewidth]{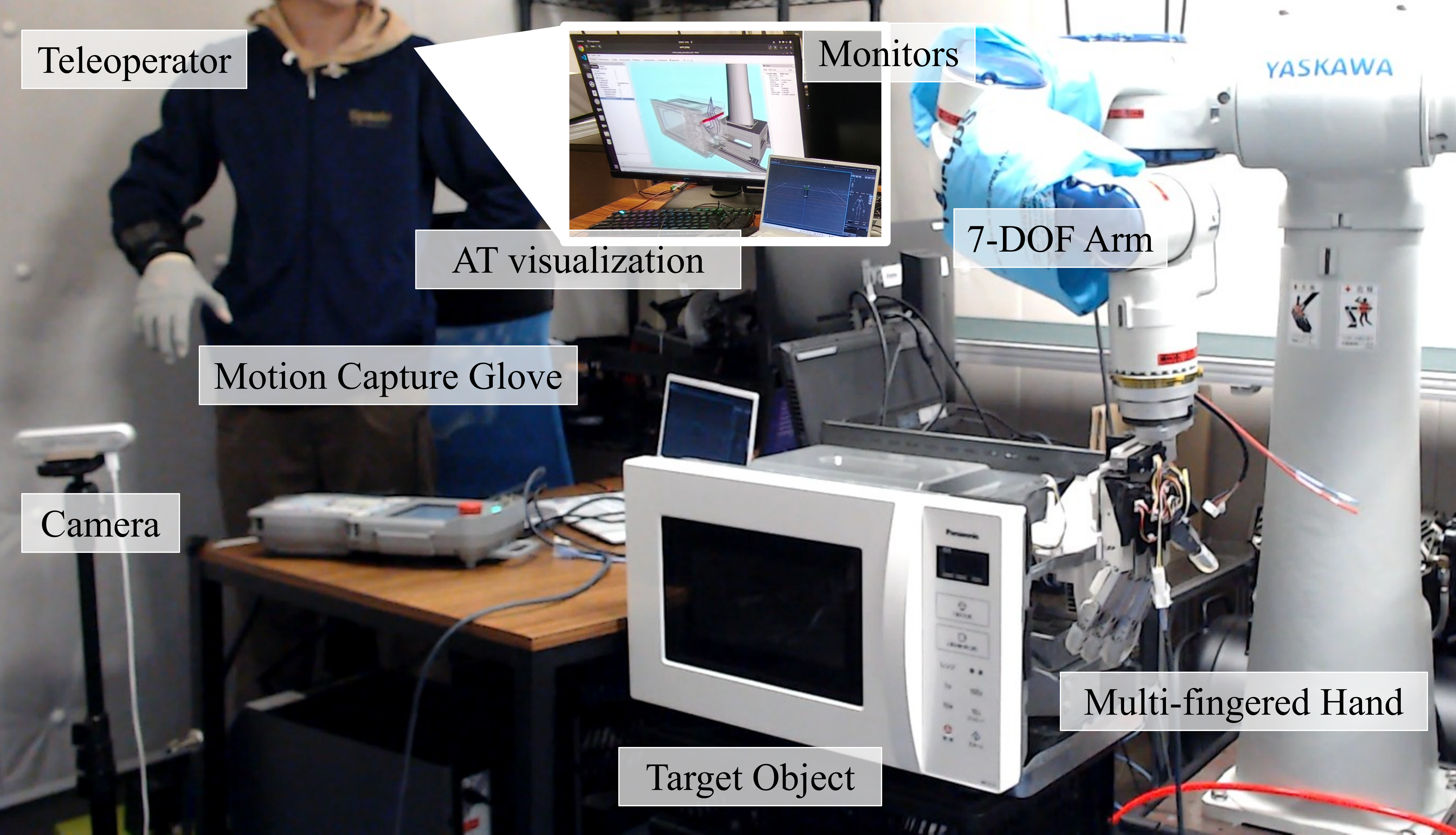}
        \subcaption{System Architecture}
    \end{minipage}
    \begin{minipage}[tb]{\linewidth}
        \centering
        \includegraphics[keepaspectratio, width=\linewidth]{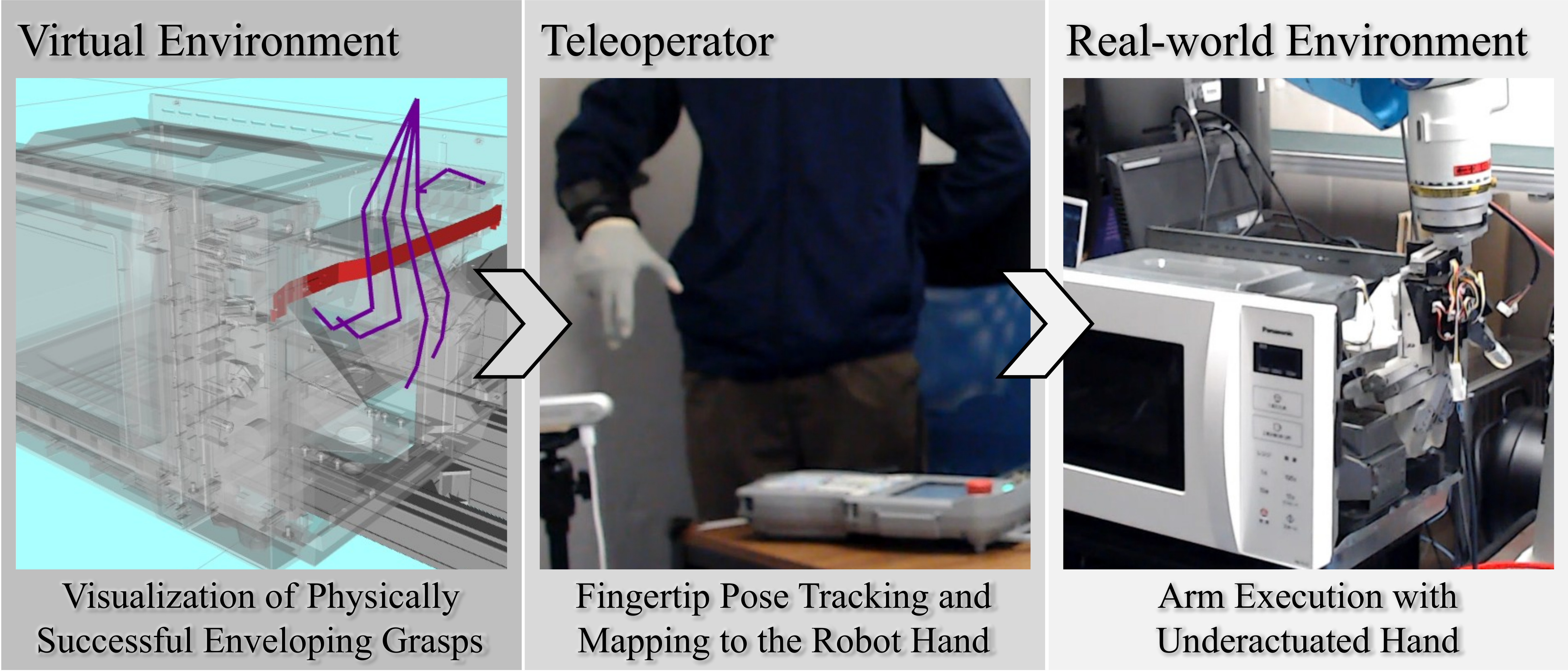}
        \subcaption{Teleoperation Procedure}
    \end{minipage}
    \caption{Overview of the proposed affordance-guided teleoperation framework. (a) illustrates the system mapping human hand motions to the underactuated robot hand. (b) outlines the procedure: physics-based AT generation, visual presentation, mapping of tracked human motions, and physical execution.}
    \figlab{overview}
\end{figure}
Affordance is a concept that describes the possibilities for action that an object provides to an agent. To embed task-relevant action possibilities (affordances) into 3D models, Hart~\etal~\cite{Hart2015} proposed the Affordance Templates (AT) framework. We utilize this framework for multi-fingered enveloping grasps. \figref{overview}~(a) shows our teleoperation system mapping human motions to a 7-DOF arm and a five-fingered hand for intuitive, AT-guided grasping. \figref{overview}~(b) illustrates our comprehensive procedure, encompassing the exploration of grasp configurations in a physics simulator, their presentation as color-coded 3D visual templates, and the mapping of tracked human fingertip poses to the underactuated robot hand for physical arm execution.

Our primary contribution is an affordance-guided enveloping grasp teleoperation framework specifically designed for the non-destructive disassembly of pinch-infeasible parts. By pre-calculating and visualizing valid enveloping grasps, our method transforms the cognitively demanding task of extracting tightly mated components into an intuitive alignment procedure.

In our experiments, we first generate ATs for multiple microwave components in physics simulation to identify robust grasp configurations. We then conduct real-robot teleoperation trials to evaluate the effectiveness of the proposed framework. By executing physical extractions of components characterized by high-friction mating structures, we demonstrate that visually presented ATs successfully guide operators even under severe visual occlusions and restricted accessibility within the appliance housing, where teleoperation relying solely on 2D camera feeds typically fails.

\section{Related Work}

\subsection{Affordance-Guided Manipulation and Teaching}
Hart~\etal~\cite{Hart2015} established the AT framework to embed task constraints into 3D models, significantly reducing the input burden in teleoperation. This concept has since been extended to mobile manipulation~\cite{James2015} and formalized into computationally efficient Affordance Primitives (AP)~\cite{Pettinger2020,Pettinger2022}, which have been successfully applied to humanoid teleoperation via Mixed Reality~\cite{Pohl2020,Penco2024}. In the context of disassembly, affordance-based methods have improved task success rates in environments cluttered with overlapping scrap components~\cite{Gorjup2019} and simplified the teaching of mating-component extraction involving rigid-link constraints~\cite{Sako2025}. Other approaches extract abstract affordances directly from human demonstrations or videos~\cite{Regal2023,Fang2018}. Recently, the cognitive benefits of ATs have also been validated in high-workload teleoperation scenarios requiring the simultaneous coordination of bimanual manipulation and mobile base navigation~\cite{Frering2024}.

While existing studies primarily rely on manually predefined affordances or restrict their application to simplified arm trajectories and parallel grasping, manually defining stable multi-point contact configurations for complex 3D geometries is highly impractical. In contrast, we systematically pre-generate ATs for multi-fingered enveloping grasps based on geometric and physical stability evaluations within a physics simulator. This physics-based automated generation eliminates the need for manual template authoring and enables us to effectively apply affordance guidance to complex, real-world non-destructive disassembly tasks using an underactuated hand.

\subsection{Teleoperation Interfaces for Multi-Fingered Hands}
Recent advancements in multi-fingered teleoperation have significantly improved how human motions are mapped to robotic hands. Studies have explored kinematic compensation~\cite{Gioioso2013}, vision-based markerless tracking~\cite{Handa2020,Qin2023}, and accessible interfaces using commercial sensors~\cite{Coppola2022}. However, these approaches primarily focus on improving input mechanisms to accurately capture human hand poses. They critically lack advanced teaching interfaces that visually guide the operator on \textit{how} and \textit{where} to grasp complex objects in unstructured environments.

In the context of appliance disassembly, this lack of visual guidance becomes a critical limitation. Because target components are deeply embedded, conventional teleoperation relying solely on raw 2D camera feeds makes teaching precise enveloping grasps practically impossible due to the loss of depth perception and severe visual occlusions. While some studies attempt to mitigate insufficient visual information through compliance control or force feedback~\cite{Ajoudani2012,Peternel2015,Talasaz2017}, safely guiding high-DOF fingers into narrow spaces fundamentally requires proactive visual augmentation. To overcome this, our system uniquely integrates an affordance rendering interface with reliable glove-based tracking, empowering the operator to precisely align and execute multi-fingered enveloping grasps even for visually occluded mating parts.

\section{Proposed Method}
\subsection{Overview}
\figref{overview}~(a) illustrates our complete teleoperation system, while \figref{objects} shows the pinch-infeasible components targeted in this study. Fingertip grasping alone cannot securely extract these tightly mated components due to the lack of graspable protrusions. Specifically, Object A requires extraction from a narrow lateral slot; Object B presents a wide, flat profile with fragile plastic edges; and Object C is a heavy, box-like metallic component located deep within the housing.
\begin{figure}[tb]
    \centering
    \begin{minipage}[tb]{0.32\linewidth}
        \centering
        \includegraphics[keepaspectratio, width=\linewidth]{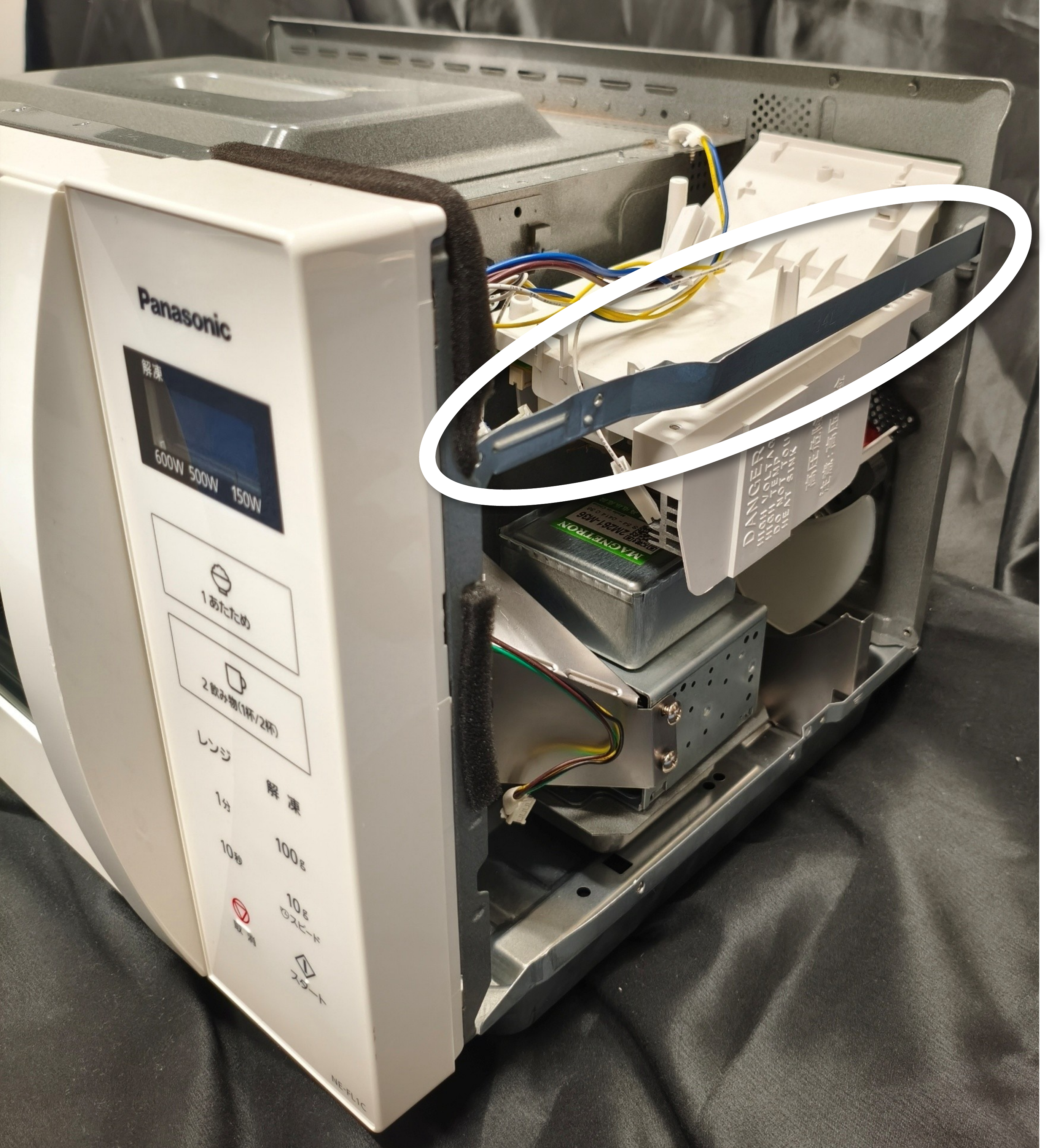}
        \subcaption{Object A}
    \end{minipage}
    \begin{minipage}[tb]{0.32\linewidth}
        \centering
        \includegraphics[keepaspectratio, width=\linewidth]{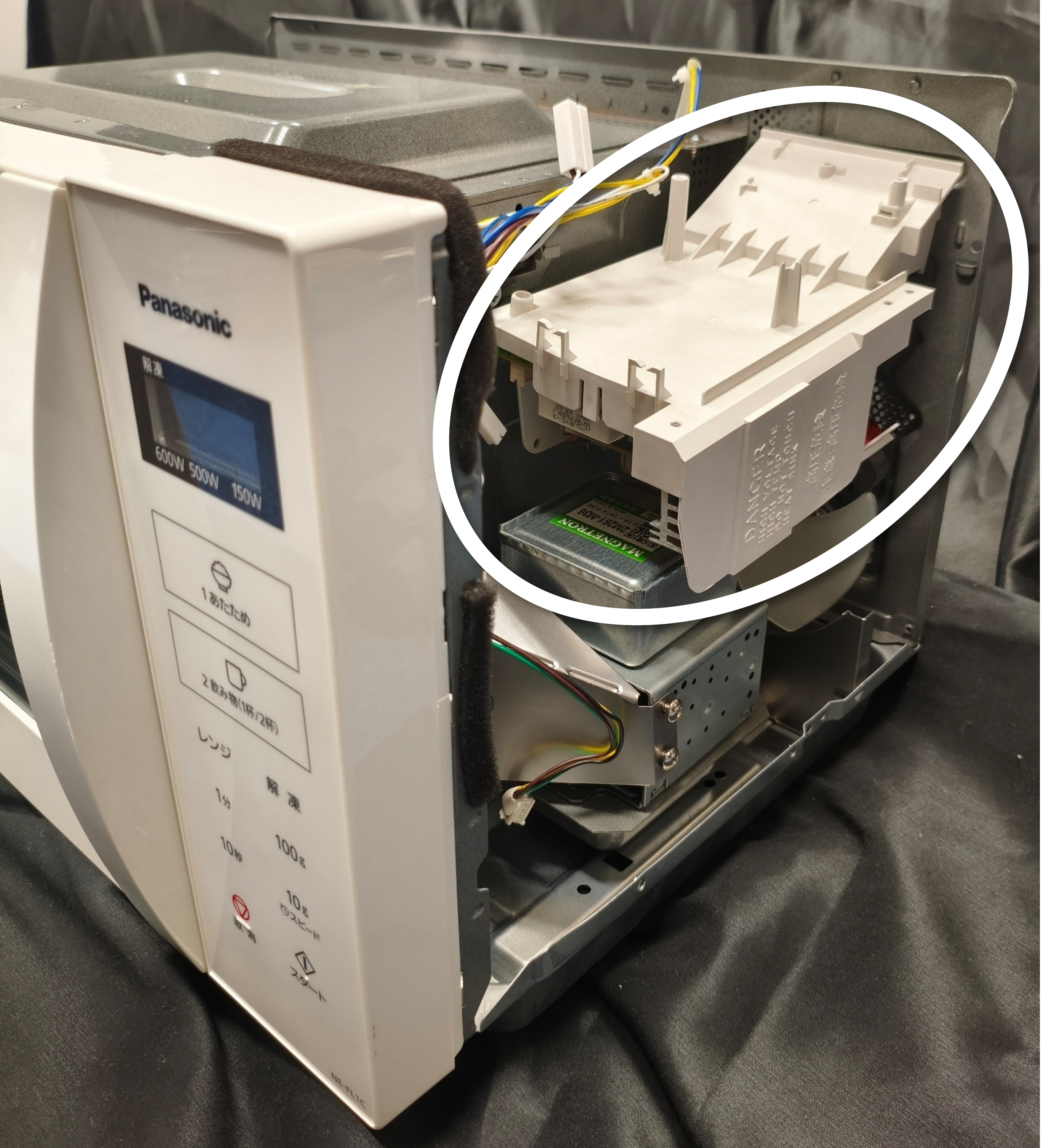}
        \subcaption{Object B}
    \end{minipage}
    \begin{minipage}[tb]{0.32\linewidth}
        \centering
        \includegraphics[keepaspectratio, width=\linewidth]{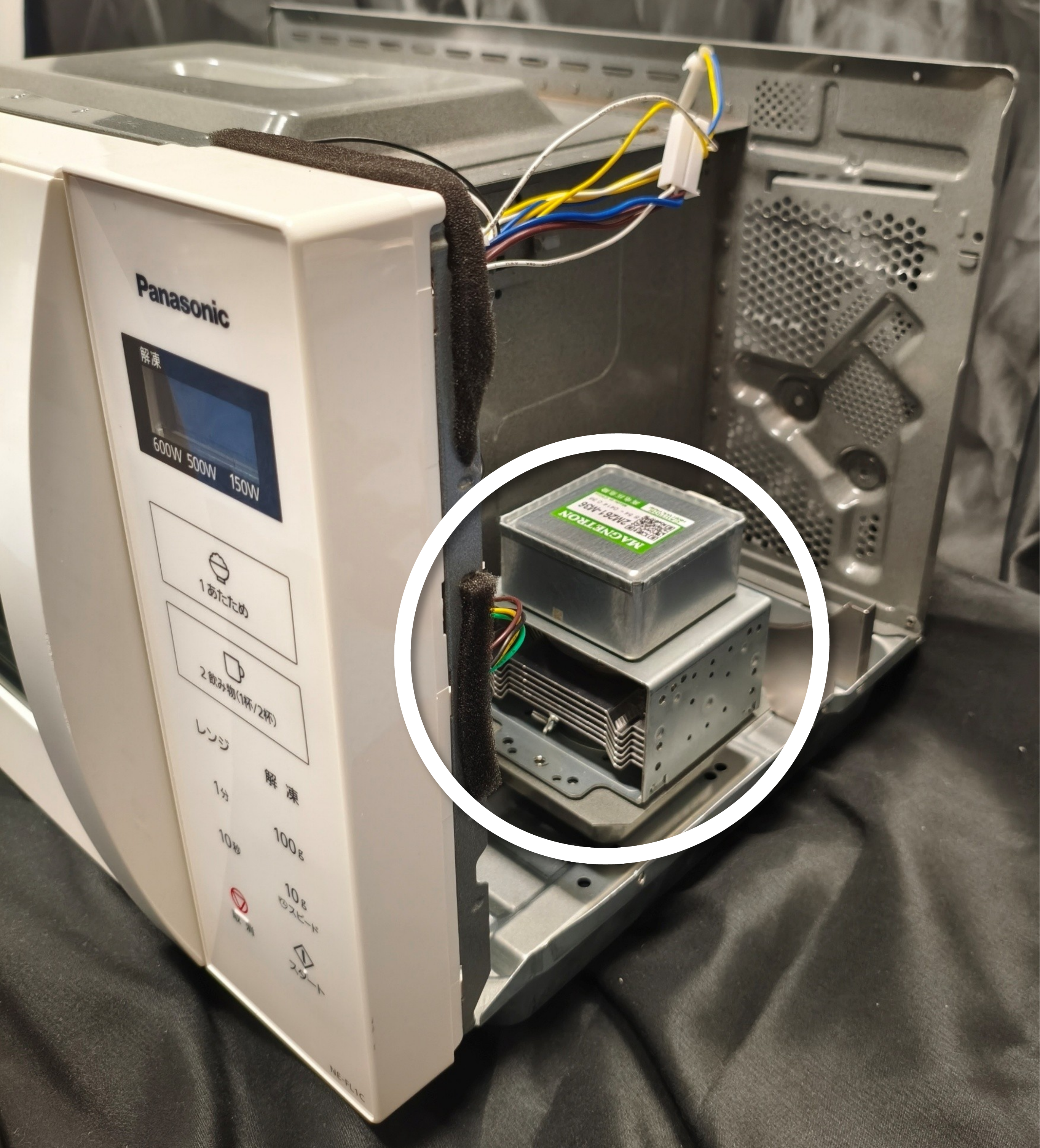}
        \subcaption{Object C}
    \end{minipage}
    \caption{Pinch-infeasible components to be disassembled. (a) Object A requires extraction from a narrow lateral slot. (b) Object B has a wide, flat profile with fragile edges. (c) Object C is a heavy, box-like metallic component located deep within the housing.}
    \figlab{objects}
\end{figure}

To overcome the strong friction and geometric constraints of these mating structures, it is mechanically essential to employ enveloping grasps that distribute contact points across multiple fingers, thereby transmitting the necessary extraction force and moments without damaging the parts. To address this, our framework (\figref{overview}~(b)) consists of three stages: 1) physics-based AT generation, 2) AT visualization for teaching, and 3) real-world execution. The operator inputs arm and finger motions visually guided by the generated ATs. The system then maps these motions to the five-fingered hand, thereby clarifying the grasp strategy in advance and significantly reducing trial-and-error during physical disassembly.

\subsection{Physics-Based Generation of ATs}
To abstract successful enveloping grasps as ATs, we employ physics simulations where hand poses are iteratively generated to enclose the target object. \figref{simulation} details this sequential generation process. First, a valid sampling region for initial hand placement is defined to avoid premature collisions with the appliance housing. The hand model is initially positioned such that its palm (base link) contacts the target object’s surface. From this state, the simulator independently actuates each finger until it collides with the object or environment, establishing a multi-point contact state including the palm.

At grasp completion, we construct a 3D convex hull $\text{Hull}(\mathcal{P})$ from the set of representative points $\mathcal{P}=\{\mathbf{p}_1,\dots,\mathbf{p}_N\}$, where $\mathbf{p}_i \in \mathbb{R}^3$ are the geometric centers of all hand links (palm and finger segments) currently in contact with or surrounding the object. Following prior studies on caging grasps~\cite{Diankov2008,Rodriguez2012}, grasp success requires the object’s center of mass $\mathbf{g}\in\mathbb{R}^3$ to reside within $\text{Hull}(\mathcal{P})$:
\begin{equation} \forlab{1}
\sum_{i=1}^N \alpha_i \mathbf{p}_i = \mathbf{g} \quad \text{s.t.} \quad \alpha_i \ge 0, \; \sum_{i=1}^N \alpha_i = 1
\end{equation}
where $\boldsymbol{\alpha} = [\alpha_1, \dots, \alpha_N]^\top$ is the weight vector. This condition ensures that the multi-fingered enclosure restricts the object's movement, preventing it from escaping during the frictional disassembly process.

To evaluate grasp quality, we calculate the distance $d_h$ between the object's center of mass $\mathbf{g}$ and the centroid of the hull vertices $\mathbf{c}$:
\begin{equation} \forlab{2}
d_h = \| \mathbf{c} - \mathbf{g} \|
\end{equation}
where $\mathbf{c} = \frac{1}{M} \sum_{j=1}^M \mathbf{v}_j$, with $M$ hull vertices $\mathbf{v}_j$. Minimizing $d_h$ implies that the geometric center of the grasp is aligned with the object's mass center. This spatial symmetry promotes an even distribution of extraction forces across the fingers and the palm, enhancing the overall robustness of the enveloping grasp against external perturbations.

\begin{figure}[tb]
  \centering
  \includegraphics[width=\linewidth]{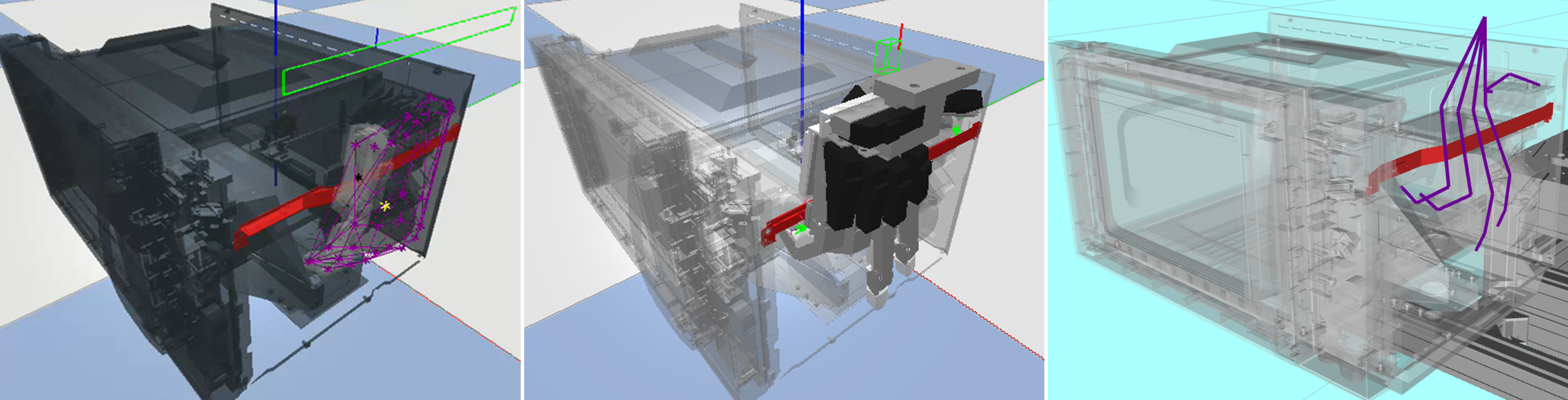}
  \caption{Sequential process of AT generation in the physics simulator. The left shows the valid sampling region and point cloud evaluation, the center displays the rigid-body interaction during grasp execution, and the right presents the successfully extracted affordance template.}
  \figlab{simulation}
\end{figure}

\subsection{AT Visualization and Coordinate Alignment}
The simulator presents the generated AT candidates to the operator in a 3D visualization environment. \figref{at_result} illustrates how these templates are color-coded based on the normalized $d_h$; smaller values representing higher grasp quality are mapped to red, while larger values are mapped to blue. This visualization allows operators to intuitively compare and select optimal grasp strategies.

To accurately reflect the simulated grasps in the real world, we align the coordinate frames using a mapping transform $\mathbf{T}_{\text{map}} = \mathbf{T}_{\text{base}}^{\text{vis}} (\mathbf{T}_{\text{base}}^{\text{sim}})^{-1}$. Here, $\mathbf{T}_{\text{base}}^{\text{sim}}, \mathbf{T}_{\text{base}}^{\text{vis}} \in SE(3)$ are the reference base poses in the physics simulator and the visualization system, respectively. An arbitrary body pose $\mathbf{T}_{\text{body}}^{\text{sim}}$ is subsequently mapped as $\mathbf{T}_{\text{body}}^{\text{vis}} = \mathbf{T}_{\text{map}} \mathbf{T}_{\text{body}}^{\text{sim}}$.

\begin{figure}[tb]
  \centering
  \includegraphics[width=\linewidth]{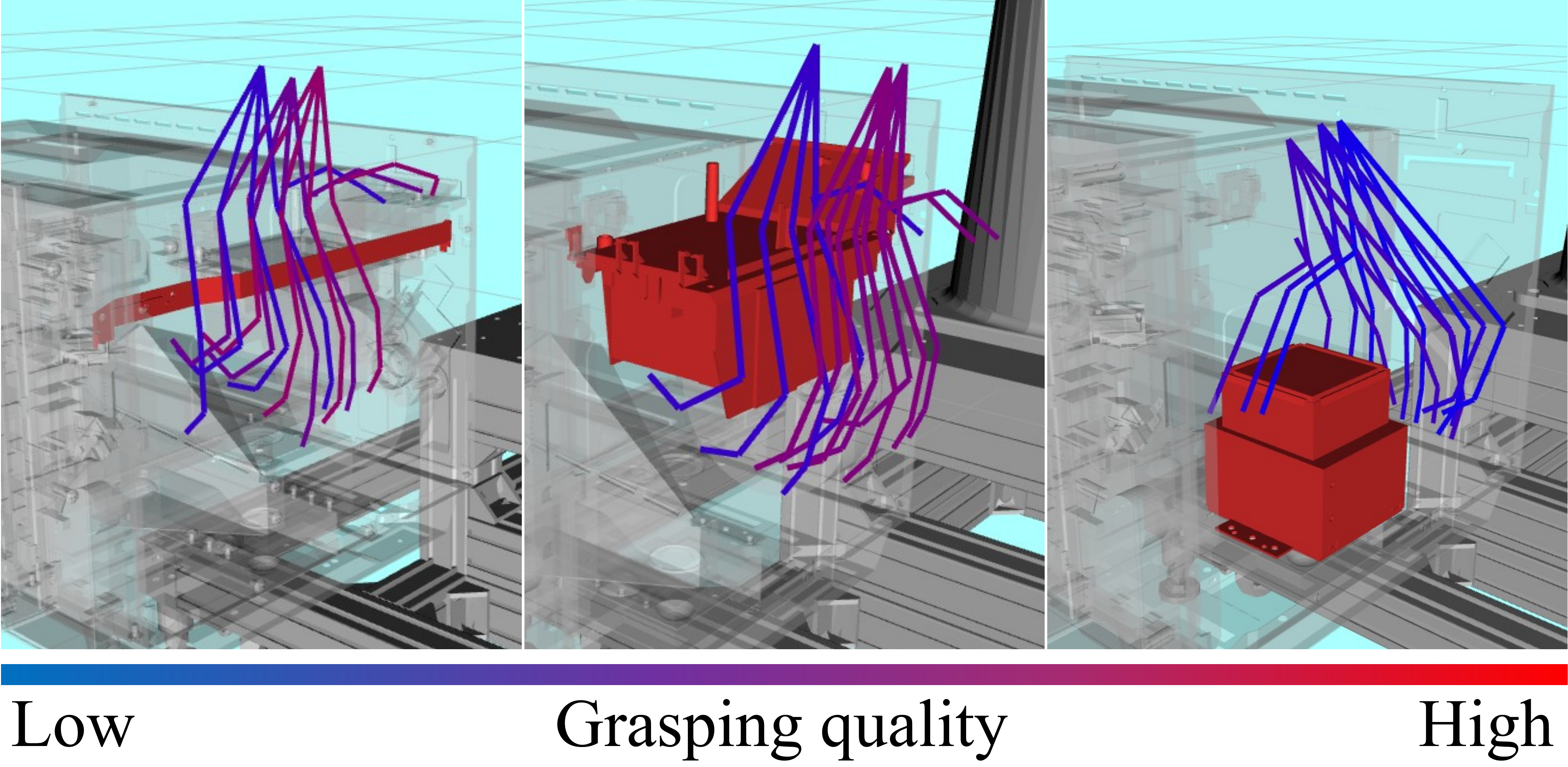}
  \caption{Generated affordance templates for target objects A, B, and C. High quality ($d_h$: Small) grasps are rendered in red.}
  \figlab{at_result}
\end{figure}

\begin{figure}[tb]
    \centering
    \begin{minipage}[tb]{0.326\linewidth}
        \centering
        \includegraphics[keepaspectratio, width=\linewidth]{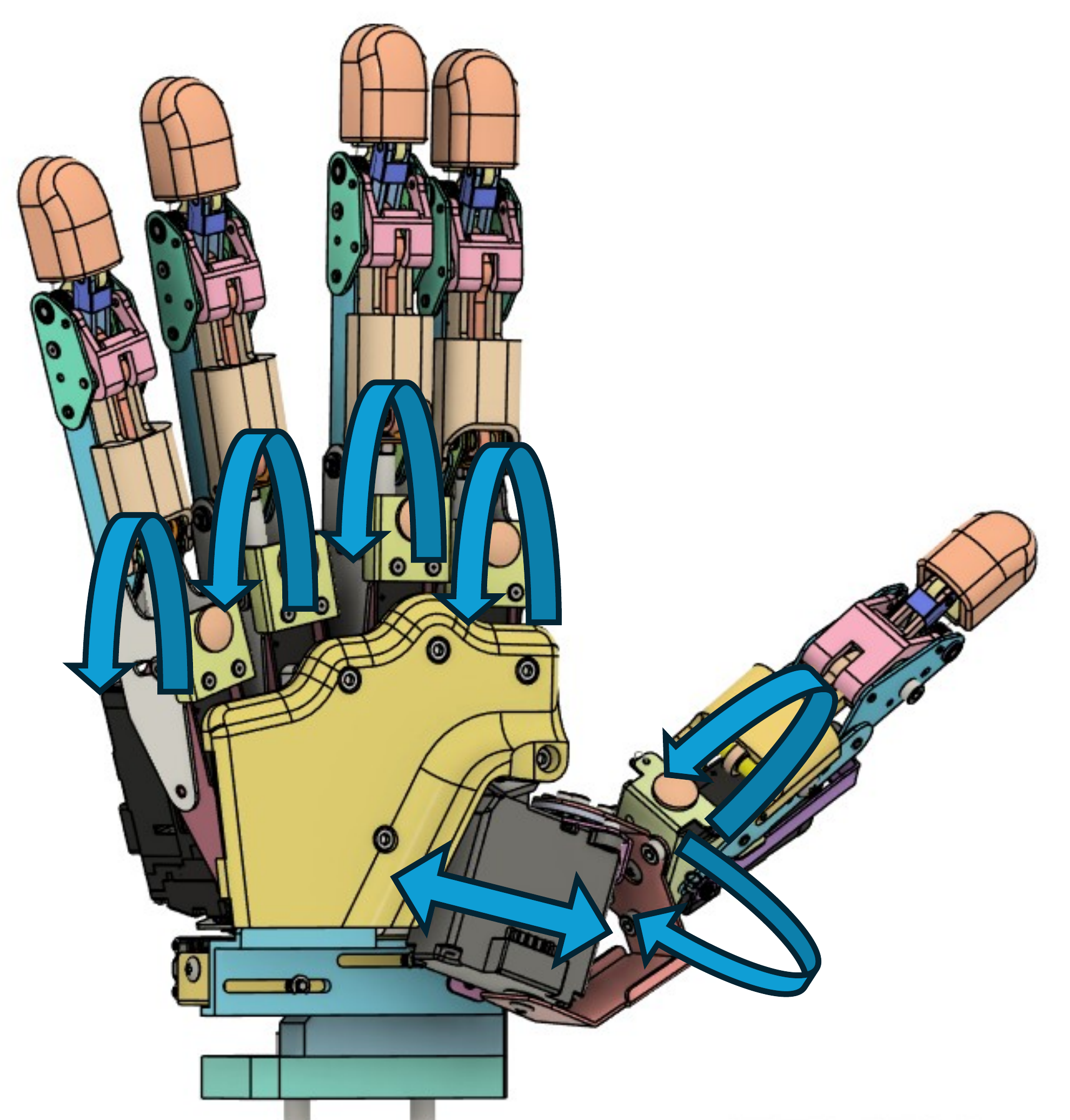}
        \subcaption{Model}
    \end{minipage}
    \begin{minipage}[tb]{0.298\linewidth}
        \centering
        \includegraphics[keepaspectratio, width=\linewidth]{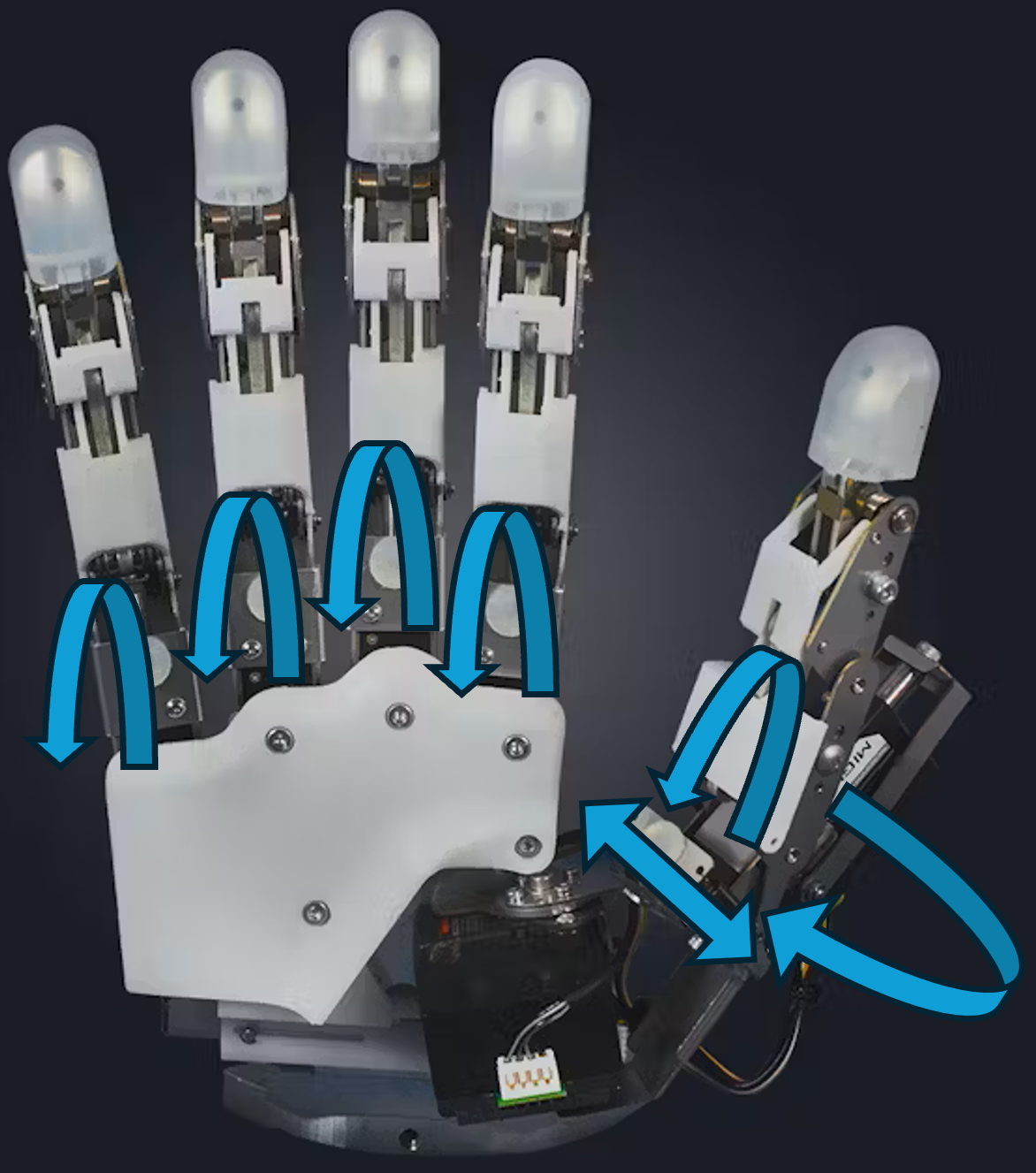}
        \subcaption{Robot hand}
    \end{minipage}
    \begin{minipage}[tb]{0.34\linewidth}
        \centering
        \includegraphics[keepaspectratio, width=\linewidth]{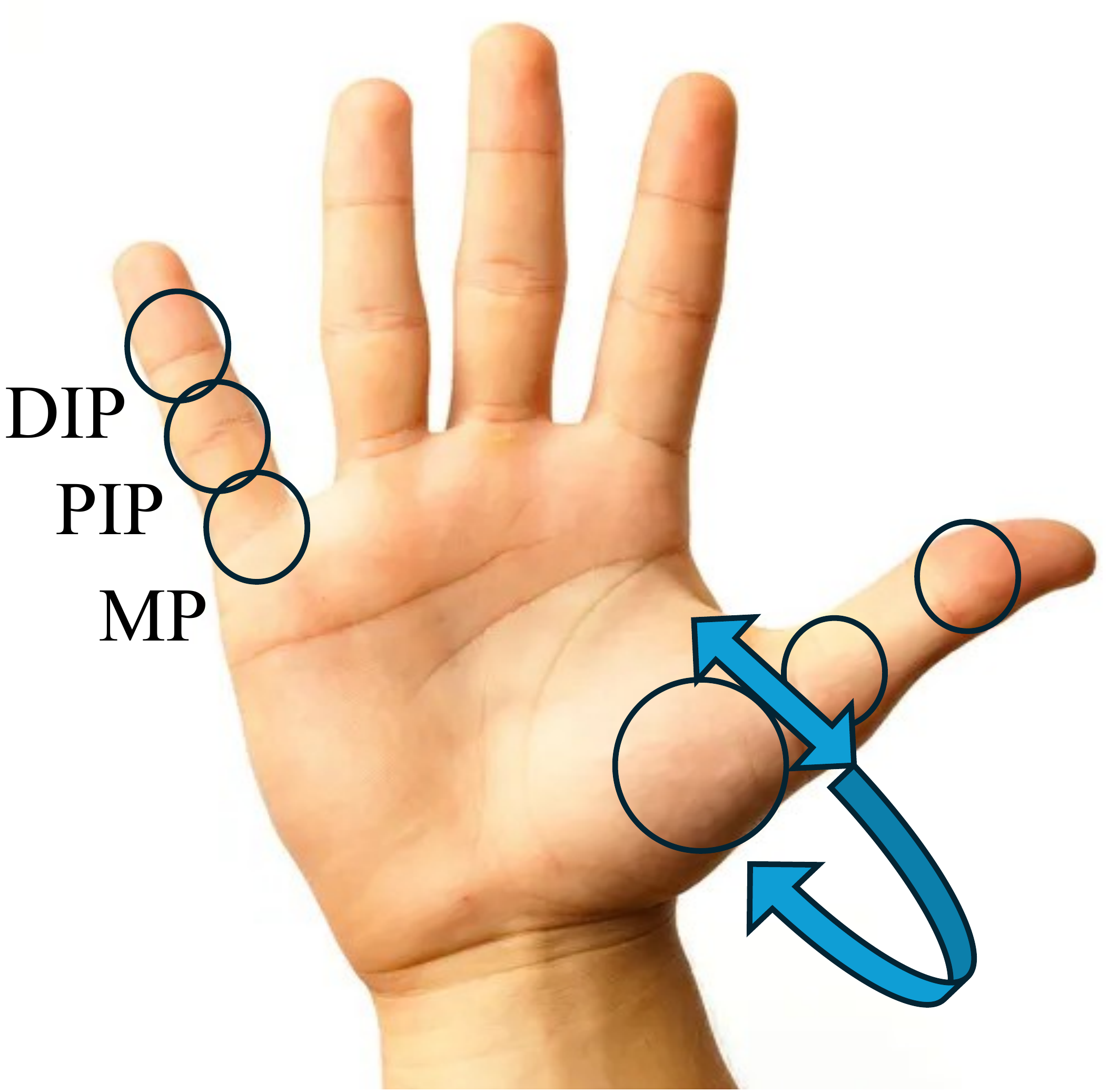}
        \subcaption{Human hand}
    \end{minipage}
    \caption{The humanoid multi-fingered hand capable of enveloping grasps and comparison with a human hand.}
    \figlab{5st}
\end{figure}

\subsection{Motion Mapping to the Underactuated Hand}
We seamlessly map human finger motions, precisely measured by Neuron Studio Gloves developed by Perception Neuron, to the underactuated robotic hand. \figref{5st} (a), (b), and (c) respectively illustrate the anatomical correspondence between the rigid-body model, the actual robotic structure, and the human hand. The human hand possesses over 20 degrees of freedom, whereas the industrial robot hand relies on only seven actuation pulses: two for thumb yaw/roll, and five for individual finger flexion. Such underactuated mechanisms excel at conforming to unknown object shapes but introduce significant kinematic dissimilarities compared to the human hand.
Let the finger set be $\mathcal{F}=\{\text{Thumb}, \text{Index}, \text{Middle}, \text{Ring}, \text{Pinky}\}$. For each finger $f \in \mathcal{F}$, the human joint angles are defined as $\boldsymbol{\theta}_h^{(f)} \in \mathbb{R}^3$. The control input vector $\mathbf{u} \in \mathbb{Z}^7$ is defined as:
\begin{equation} \forlab{3}
\mathbf{u} = [u_{\text{Th}}^{\text{flx}}, u_{\text{Idx}}^{\text{flx}}, u_{\text{Mid}}^{\text{flx}}, u_{\text{Rng}}^{\text{flx}}, u_{\text{Pky}}^{\text{flx}}, u_{\text{Th}}^{\text{yaw}}, u_{\text{Th}}^{\text{rol}}]^\top
\end{equation}
where subscripts represent abbreviated finger names (e.g., Idx for Index) and superscripts denote motion types (flexion, yaw, roll).

Because direct joint-to-joint mapping is kinematically impossible due to the underactuated mechanism---where a single motor drives multiple coupled joints simultaneously---the flexion pulse $p_f$ is computed using predefined lookup tables $\mathcal{T}_f=\{(p_k, \boldsymbol{\Theta}_k)\}_{k=1}^K$. These tables are constructed through prior human calibration measurements, capturing the nonlinear curling trajectory of each robotic finger. Here, $K$ is the total number of recorded postures, and $\boldsymbol{\Theta}_k$ is the human posture corresponding to pulse $p_k$. We select the optimal index $k^*$ that minimizes the Euclidean distance between the current human posture $\boldsymbol{\theta}_h^{(f)}$ and the recorded posture $\boldsymbol{\Theta}_k$:
\begin{equation} \forlab{4}
k^* = \arg\min_{1 \le k \le K} \| \boldsymbol{\theta}_h^{(f)} - \boldsymbol{\Theta}_k \|
\end{equation}

\section{Experiments}

\subsection{Experimental Setup}
We designed our experiments to validate AT generation in simulation and to verify teleoperation performance on a real robot. \figref{objects} shows the disassembly targets, which consist of three pinch-infeasible microwave components (Objects A, B, and C). These components possess complex mating structures and lack clear protrusions for fingertip grasping.

The physics simulator automatically generates dynamic grasp trials and extracts representative successful grasps based on the $d_h$ quality metric. The physical setup uses a YASKAWA Motoman SDA5F 7-DOF arm equipped with a Double Giken D-Hand 5ST. The operator intuitively controls the multi-fingered hand using Neuron Studio Gloves, which provide high-frequency (approx. 120 Hz) IMU-based finger joint tracking. Simultaneously, arm spatial motions are synchronized via MediaPipe~\cite{Lugaresi2019}-based wrist tracking, which robustly estimates the operator's 6D wrist pose from a monocular RGB camera operating at 30 fps. The spatial coordinate transformation between the human workspace and the robot workspace is carefully calibrated to ensure that the operator's arm movements seamlessly align with the visually superimposed ATs on the monitor. The entire control loop, including the nearest-neighbor posture mapping defined in~\forref{4}, operates with a latency of less than 50 ms, providing a highly responsive teleoperation experience.

\subsection{Simulation Results of AT Generation}
\figref{at_result} shows the generated ATs for Objects A, B, and C. The proposed method successfully extracted a diverse set of representative enveloping grasps for the different mating components, varying significantly in approach direction and finger aperture. For example, Object B's templates suggest a wider thumb placement to span its broad width, while Object C's templates indicate deeply enveloping configurations that maximize contact area along the sides of the heavy, box-like structure. This result demonstrates the framework’s generality across varying geometries without relying on manually predefined templates.

\subsection{Real-World Teleoperation Results}
To evaluate physical reproducibility and disassembly success, the operator performed teleoperation tasks guided by the visual ATs. For Object A, we selected three distinct ATs categorized by their quality metrics (High, Medium, and Low scores) and conducted 10 independent extraction trials for each configuration. \figref{real_exp_result} provides sequential snapshots of the physical execution for Object A. The process typically unfolds in four phases: (1) approaching the target via spatial arm tracking, (2) fine-tuning the wrist orientation to align with the visual AT overlay, (3) closing the fingers to achieve the enveloping grasp, and (4) executing the final extraction motion.

\tabref{real_results} summarizes the success rates: 90\% for the High-score AT, 100\% for the Medium-score AT, and 80\% for the Low-score AT. Failures typically occurred when the fingertips insufficiently engaged the object’s surfaces prior to the pulling motion, causing the hand to slip. Specifically, failures in the Low-score AT occurred because the initial contact points were located too far from the center of mass $\mathbf{g}$ (\ie, corresponding to a large $d_h$). This spatial misalignment generated an unintended large moment during the extraction phase, breaking the caging condition defined in~\forref{1} and causing the object to rotate and slip out of the robotic fingers.

Furthermore, \figref{executed_grasps} demonstrates that the unique AT postures generated for all three target objects were physically reproducible. The operator effectively utilized the visual ATs to modulate the finger apertures to match the varying geometric demands of each object. As seen in \figref{executed_grasps}~(b), the operator executed a wide thumb spread to securely hold the broad profile of Object B, while \figref{executed_grasps}~(c) illustrates the deep, hook-like enveloping strategy required to securely extract the heavy Object C from its deep housing.

\begin{figure*}[tb]
  \centering
  \includegraphics[width=\linewidth]{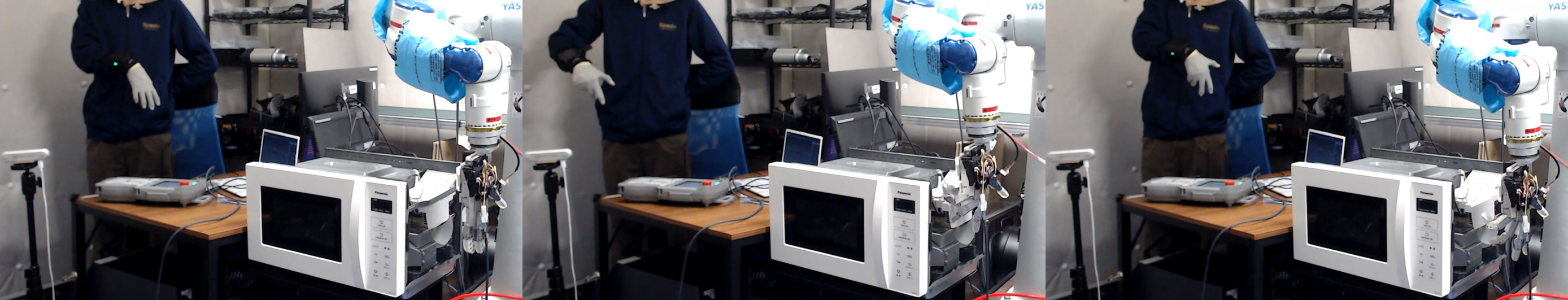}
  \caption{Sequential snapshots of the real-world disassembly process for Object A. The sequence illustrates three key stages: (1) approaching the target via spatial wrist tracking, (2) executing the enveloping grasp guided by the visual AT overlay, and (3) successfully extracting the mating component.}
  \figlab{real_exp_result}
\end{figure*}

\begin{table}[tb]
  \centering
  \caption{Success rates of real-world disassembly trials for Object A (10 independent trials per configuration).}
  \tablab{real_results}
  \setlength{\tabcolsep}{5mm}
  \begin{tabular}{lr}
    \toprule
    \multicolumn{1}{c}{AT Quality Level} & \multicolumn{1}{c}{Success Rate [\%]} \\
    \midrule
    High ($d_h$: Small) & 90 \\
    Medium ($d_h$: Mid) & 100 \\
    Low ($d_h$: Large) & 80 \\
    \bottomrule
  \end{tabular}
\end{table}

\begin{figure*}[tb]
    \centering
    \begin{minipage}[tb]{0.32\linewidth}
        \centering
        \includegraphics[keepaspectratio, width=\linewidth]{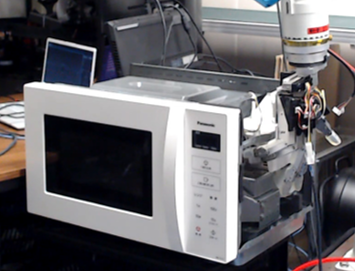}
        \subcaption{Object A}
    \end{minipage}
    \begin{minipage}[tb]{0.32\linewidth}
        \centering
        \includegraphics[keepaspectratio, width=\linewidth]{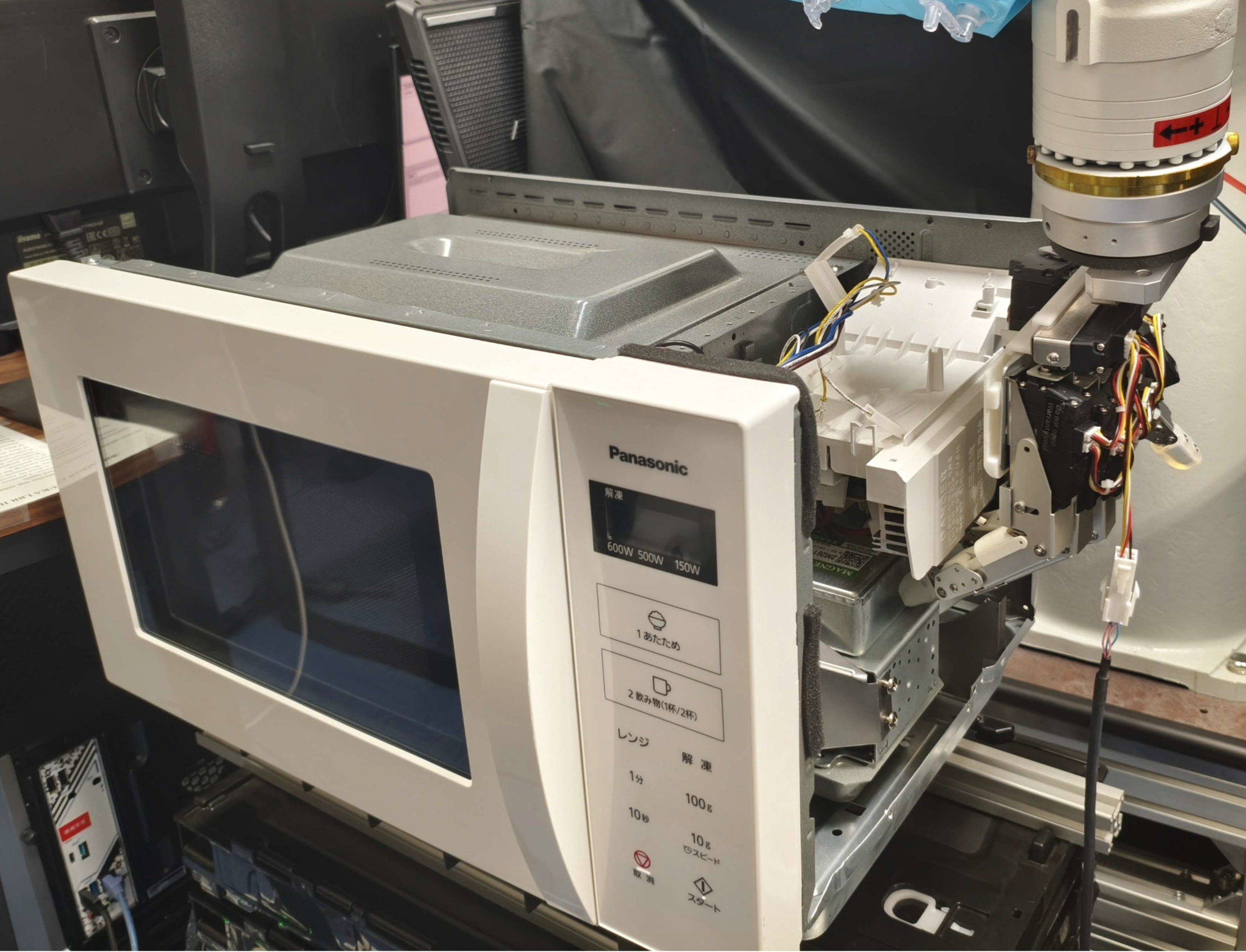}
        \subcaption{Object B}
    \end{minipage}
    \begin{minipage}[tb]{0.32\linewidth}
        \centering
        \includegraphics[keepaspectratio, width=\linewidth]{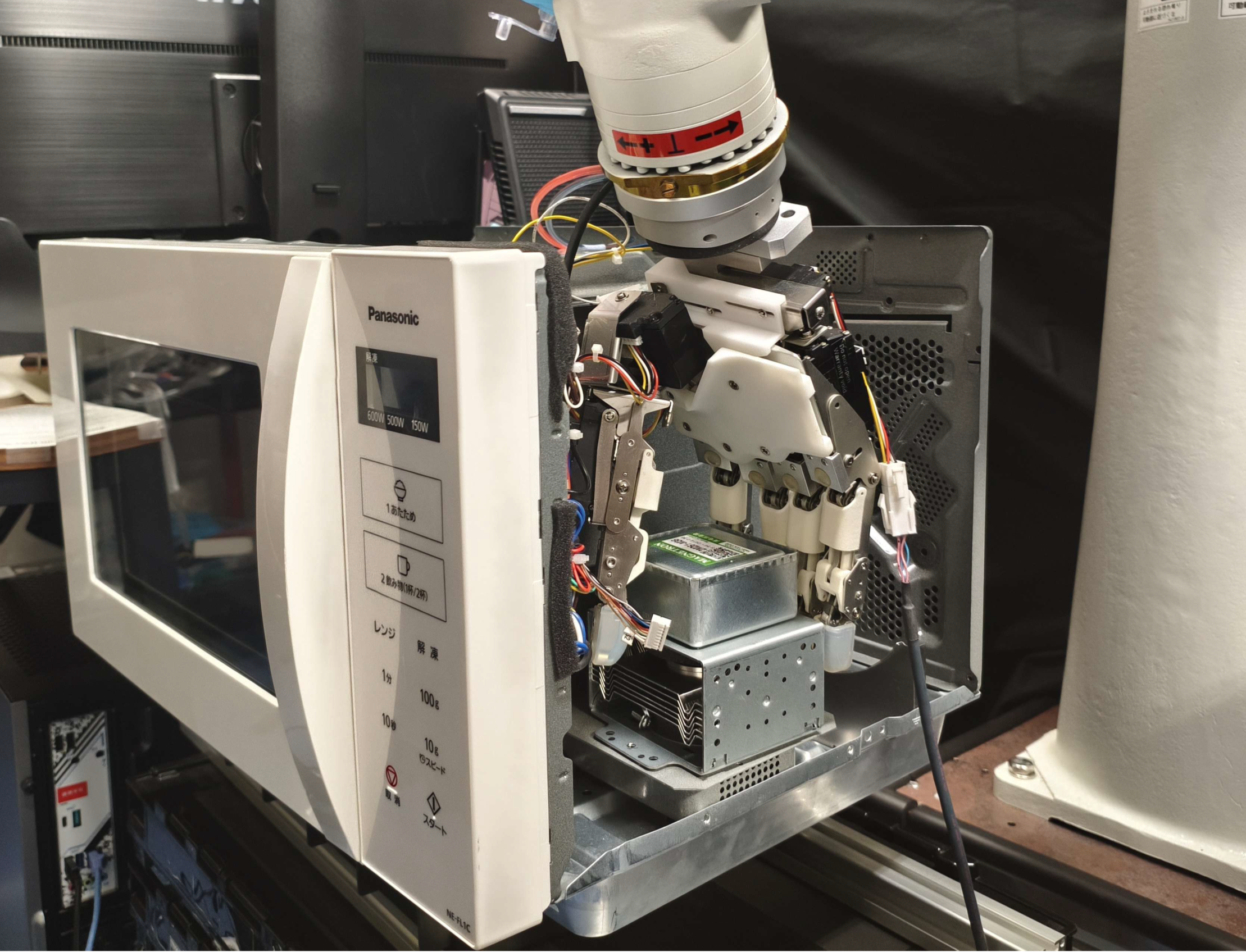}
        \subcaption{Object C}
    \end{minipage}
    \caption{Executed enveloping grasps guided by ATs. The operator successfully modulated finger apertures to match varying geometric demands: (a) a lateral grasp for Object A, (b) a wide thumb spread for the broad profile of Object B, and (c) a deep, hook-like enveloping strategy to securely extract the heavy Object C.}
    \figlab{executed_grasps}
\end{figure*}

\section{Discussion}

\subsection{Visual Augmentation and the Rationale for Multiple Candidates}
Because target components are embedded deep within the appliance casing, raw visual information alone fails to capture the required depth and contact points. The visually superimposed AT candidates act as a spatial guide, effectively turning a cognitively demanding blind grasping task into an intuitive 6D pose alignment procedure. This eliminates the need for continuous mental translation between 2D and 3D environments, reducing the operator's mental workload, consistent with mixed-reality teleoperation studies~\cite{Zolotas2019}.

Interestingly, \tabref{real_results} shows a 100\% success rate for the Medium-score AT compared to the 90\% rate for the High-score AT. While analytic grasp quality measures are theoretically rigorous, they are notoriously sensitive to shape uncertainties~\cite{Roa2015}. A mathematically perfect "High-score" grasp tightly fits the ideal CAD model but may lack robustness against physical noise, calibration drift, or minor geometric discrepancies. In contrast, the Medium-score AT inherently provides slightly more clearance and structural robustness. This practical discrepancy justifies our core design choice: providing discrete, robust options and empowering the human to make contextual decisions yields higher system reliability than strictly enforcing autonomous computations~\cite{Losey2018}.

\subsection{Sim-to-Real Discrepancies and Teleoperation Constraints}
Despite its demonstrated effectiveness, this study has limitations. Unavoidable sim-to-real discrepancies, such as unmodeled friction and the complex kinematic responses of underactuated hands, can occasionally induce grasp instability. As analyzed by Odhner \etal~\cite{Odhner2014}, while the inherent compliance of underactuated hands aids in adapting to unknown object shapes, their exact post-contact mechanics are exceptionally difficult to predict in rigid-body simulators, inherently widening the sim-to-real gap in precision manipulation~\cite{Collins2019}.

Consequently, this highlights the necessity of incorporating a post-grasp physical verification phase (\eg, tactile sensing or visual slip tracking) into future teleoperation protocols. Furthermore, hardware safety constraints regarding the robot's payload capacity and high-friction extraction risks limited our ability to conduct full physical trials for the heavier components (Objects B and C).

\subsection{Towards Application to Robot Learning and Haptics}
Ultimately, this low-cognitive-load interface can serve as a practical foundation for collecting high-quality demonstration data for future autonomous systems. As extensively reviewed in the robot learning literature~\cite{Kroemer2021}, Imitation Learning (IL) requires large volumes of successful task executions. Historically, acquiring such data has been bottlenecked by the difficulty of multi-fingered teleoperation. By utilizing our framework, operators can efficiently generate numerous successful enveloping grasp demonstrations even for complex mating parts, directly bridging the data gap currently hindering fully autonomous non-destructive disassembly.

Furthermore, while our system relies exclusively on visual augmentation, incorporating wearable haptic feedback devices~\cite{Pacchierotti2017} represents a promising extension. Rendering simple vibrotactile cues upon simulated contact or successful AT alignment would provide redundant sensory channels. This multisensory approach would further mitigate sim-to-real discrepancies and reduce visual fatigue, enhancing overall grasp success without exclusively relying on visual attention.

\section{Conclusion} 
This study presented an affordance-guided teleoperation framework for the non-destructive disassembly of pinch-infeasible components using a multi-fingered hand. By extracting and visualizing enveloping-grasp Affordance Templates (ATs) via physics simulation, our system empowers operators to teach complex grasp postures despite severe visual occlusions. Simulation results demonstrated that the proposed method can systematically generate diverse and robust grasp configurations across varying object geometries without manual template authoring. Real-world experiments validated that human operators could effectively select and physically reproduce these pre-generated ATs, achieving high success rates in actual disassembly tasks. Future efforts will focus on bridging the sim-to-real gap regarding contact dynamics and extending the system's load capacity to further enhance the reliability of remote robotic recycling.

\section*{Acknowledgement}
This work was supported by the New Energy and Industrial Technology Development Organization (NEDO) project JPNP23002. 

\bibliographystyle{IEEEtran}
\bibliography{references}

\end{document}